\documentclass[letterpaper]{article} 
\usepackage{aaai2027}  
\nocopyright
\usepackage[hyphens]{url}  
\usepackage{graphicx} 
\usepackage{natbib}  
\usepackage{caption} 
\usepackage{algorithm}
\usepackage{algorithmic}
\usepackage{amsmath}
\usepackage{multirow}
\usepackage{bbold}
\usepackage{adjustbox}
\usepackage{newfloat}
\usepackage{listings}
\DeclareCaptionStyle{ruled}{labelfont=normalfont,labelsep=colon,strut=off} 
\floatstyle{ruled}
\newfloat{listing}{tb}{lst}{}
\floatname{listing}{Listing}

\usepackage{booktabs}

\title{Not All Visual Tokens Are Equally Safe to Remove:\\Consequence-Sensitive Visual Token Compression}
\author{
Jingbo Wen$^{1}$ \quad Liang He$^{2,\ast}$ \quad Mingyu Cao$^{3}$ \quad Haoyu Wang$^{4}$ \\[2pt]
Minxuan Hu$^{5}$ \quad Kangning Cui$^{6}$ \quad Xilu Wang$^{3}$ \\[4pt]
$^{1}$The University of Sydney \quad
$^{2}$Tongji University \quad
$^{3}$University of Surrey \quad
$^{4}$Nankai University \\[2pt]
$^{5}$Cornell University \quad
$^{6}$City University of Hong Kong \\[4pt]
\texttt{hel9919@163.com} \\[2pt]
$^\ast$Corresponding author
}
\affiliations{
    \textsuperscript{\rm }
}

\begin{document}

\maketitle

\begin{abstract}
Visual token compression for vision--language models (VLMs) has largely relied on criteria such as attention, redundancy, and uncertainty to maximize average accuracy under a fixed compute budget, implicitly assuming that all errors carry equal cost. However, the consequence of an incorrect prediction on downstream tasks is rarely symmetric: misreading an invoice amount can be far more costly than misclassifying a background color. Motivated by this, we introduce consequence-sensitive visual token compression, which allocates visual computation across requests according to their potential error costs. Our method follows a calibrate-then-allocate procedure, estimating consequence-specific error-budget curves offline and applying the calibrated token budgets online using consequence signals available from question or task information. On a controlled within-task benchmark, high- and low-consequence questions are drawn from the same document images, so content alone cannot reveal which questions are costly to get wrong. In this setting, our method reduces high-stakes errors from 0.300 to 0.133 under the same total token budget, whereas a content-driven allocator performs no better than uniform allocation. Measuring how error rates change with token budget across different cost ratios, we derive an allocation frontier: uniform allocation is optimal when errors are equally costly, and token transfer toward high-consequence questions becomes increasingly beneficial as the cost gap grows. This allocation principle generalizes well across three dense vision-language benchmarks, two budget realization mechanisms (token deletion and resolution reallocation), two VLM architectures, and multiple token selection strategies. On a realistic mixed workload, consequence-sensitive allocation reduces cost-weighted error by 38\% while achieving approximately 21\% lower latency than full-resolution inference.

\end{abstract}
\section{Introduction}

Vision tokens dominate the inference cost of modern vision--language models (VLMs), motivating a growing body of work on visual token compression. Existing approaches reduce token computation using signals derived from the input itself, including attention-based importance \citep{fastv}, feature redundancy \citep{sparsevlm,visionzip}, and layer-wise reduction schedules \citep{pyramiddrop}. These methods have achieved strong accuracy--efficiency trade-offs by optimizing a common objective, i.e., preserving average accuracy while reducing the number of processed visual tokens. This objective implicitly assumes that all errors have equal consequences on downstream tasks.

\begin{figure}[t]
\centering
\includegraphics[width=\columnwidth]{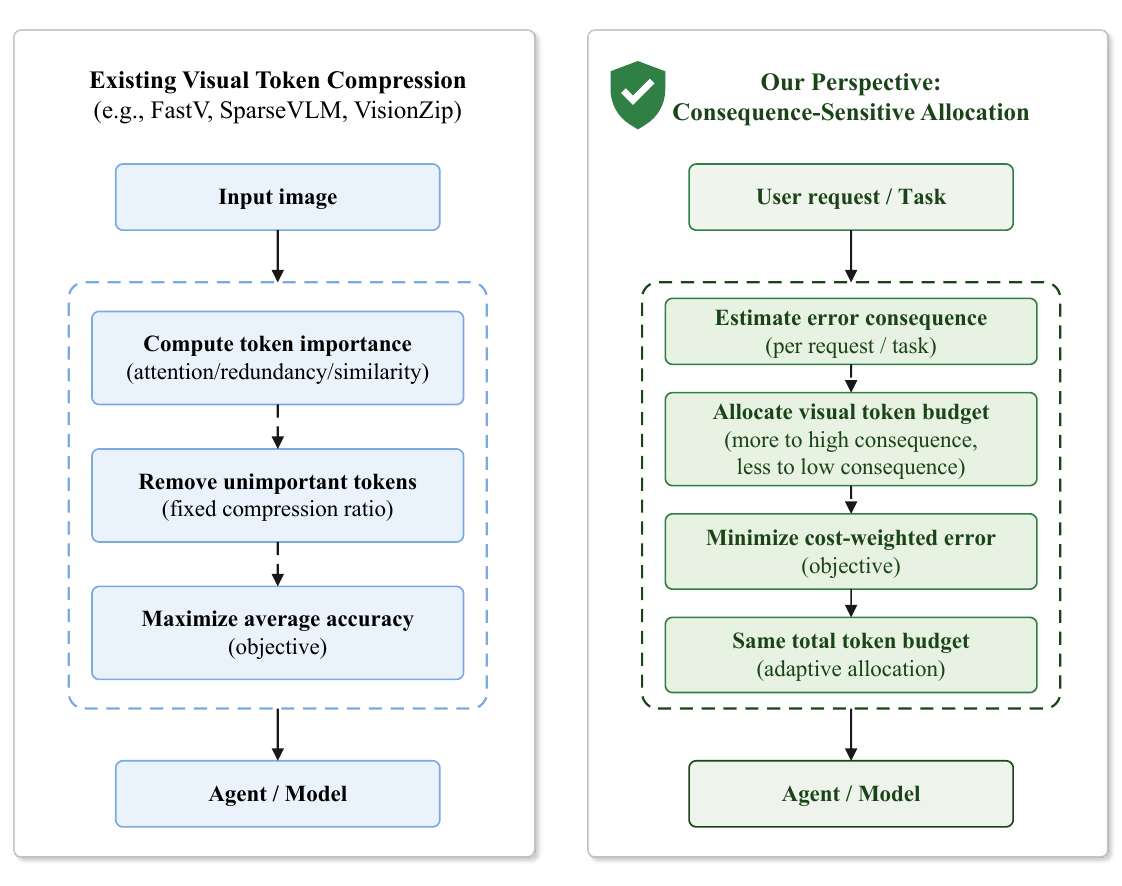}
\caption{
\textbf{Why consequence matters.}
Existing compression treats all errors equally, whereas consequence-sensitive compression prioritizes samples where mistakes are more costly.
}
\label{fig:motivation}
\end{figure}

However, real-world deployments rarely satisfy this assumption, as illustrated in Figure~\ref{fig:motivation}. In document, chart, and infographic understanding, VLM outputs often support downstream decisions and workflows. Misreading an invoice amount, transaction identifier, or medical value can trigger different consequences than misreading a section title or a decorative attribute, so the cost of an error is inherently asymmetric across samples. Importantly, such asymmetry is often available before inference through request-level information, such as question type, task identity, or application workflow. Existing visual token allocation strategies cannot exploit this signal: attention, redundancy, and uncertainty describe properties of the visual input, but do not capture the downstream consequence of being wrong.

This observation leads to a different perspective on visual token compression. Instead of asking which visual tokens are redundant, we ask where a limited visual computation budget should be allocated when different errors have different consequences. Given a set of requests with per-sample consequence weights $c_i$ and a fixed total vision-token budget, we study how to assign per-image token budgets to minimize cost-weighted error (CWE). This formulation raises three fundamental questions: 1) Does consequence information provide value beyond content-based allocation signals? To answer this, we construct a within-task benchmark where high- and low-consequence questions share the same distribution of dense document images. Since visual content is identical across consequence tiers, content-based allocators cannot exploit consequence information, allowing the effect of cost-aware allocation to be isolated. 2) When does consequence-sensitive allocation outperform uniform allocation, and how should tokens be transferred under different levels of error asymmetry? We characterize the relationship between token budgets, error reduction, and consequence ratios through measured error--budget curves. 3) Does this effect depend on a particular token reduction mechanism, model architecture, or token selector, or does it reflect a general allocation principle? We test this by varying each factor independently while holding the allocation principle fixed.

Our contributions are summarized as follows:

\begin{itemize}

\item 
We introduce consequence-sensitive visual token compression, which allocates visual computation by minimizing cost-weighted error under a strict equal total token budget. Unlike existing adaptive pruning approaches that determine budgets from content-derived signals, our new formulation incorporates inference-time-observable consequence information to guide computation allocation.

\item 
We design a within-task evaluation protocol where high- and low-consequence questions are paired with the same image distribution, eliminating content-based shortcuts. Under this setting, consequence-sensitive allocation reduces high-stakes error by $2.25\times$ under the same realized token budget.

\item 
We characterize how the optimal visual token allocation changes with error asymmetry. By analyzing measured error--budget curves, we identify a transition regime where uniform allocation is optimal under symmetric costs and token transfer toward high-consequence samples becomes increasingly beneficial as cost asymmetry grows. A calibrate-then-allocate procedure recovers near-optimal operating points on held-out data.

\item
The allocation principle generalizes across information-dense benchmarks, i.e., DocVQA, ChartQA, and InfographicVQA; two budget realization mechanisms, i.e., token deletion and resolution reallocation; two VLM architectures; and multiple token selection strategies. On a realistic mixed workload, it reduces cost-weighted error by $38\%$ while achieving substantial latency reduction compared with full-resolution inference.

\end{itemize}

\section{Related Work}
\paragraph{Visual token pruning and selection.}
Visual token compression has become an important direction for efficient VLM inference. Various methods, such as FastV \citep{fastv}, SparseVLM \citep{sparsevlm}, PyramidDrop \citep{pyramiddrop}, and VisionZip \citep{visionzip}, reduce visual computation by identifying less informative tokens using signals such as attention, redundancy, or similarity patterns. These approaches primarily address the within-image question of which tokens should be retained under a given compression ratio. In contrast, we study a complementary problem: given a global token budget across requests, how should it be distributed across samples with different error consequences? We show this allocation principle is orthogonal to token selection and holds across different selectors.
\paragraph{Adaptive visual token allocation.}
Recent methods further adapt the amount of visual computation per input. 
AdaptVision learns per-sample visual acquisition policies using reinforcement learning \citep{adaptvision}; AVIS adapts visual context and reasoning scale using redundancy and difficulty signals \citep{avis}; and PSCA and COAST adjust budgets according to information density or uncertainty-related cues \citep{psca,coast}. 
These methods allocate computation based on input properties or prediction difficulty. 
Orthogonally, consequence-sensitive allocation considers varying consequences by using inference-time-observable cost signals, such as question type or task identity, and we separate this axis from content-driven allocation in controlled within-task experiments.

\paragraph{Cost-sensitive computation.}
Cost-sensitive learning incorporates unequal error costs into prediction objectives \citep{elkan2001}. 
Related ideas appear in adaptive inference, including cost-sensitive early exit and selective prediction, where computation or coverage is adjusted according to risk and budget constraints \citep{eero}. 
Recent consequence-aware reasoning compute allocation further shows that language-model reasoning budgets can be allocated by downstream error consequence rather than difficulty alone \citep{caopd}. 
We extend this principle to VLM inference, where the controllable resource is visual token computation and consequence must be disentangled from visual redundancy and token sensitivity.

\paragraph{Vision token deletion with multimodal positional encoding.}
Practical evaluation of visual token compression requires faithful removal of visual information rather than masking or replacement. This is challenging for modern VLMs with multimodal positional encoding mechanisms, such as Qwen2.5-VL, where visual positions encode spatial information. Recent studies have analyzed positional integrity issues introduced by token pruning \citep{idalign,ppe}. We implement exact visual token deletion by preserving the original spatial coordinates of retained tokens and performing decoding with explicitly controlled embeddings and positions. This implementation enables faithful evaluation of token allocation strategies and is provided as a reproducibility artifact.
\section{Consequence-Sensitive Token Compression}
\label{sec:method}

\begin{figure}[t]  
\centering
\includegraphics[width=\columnwidth]{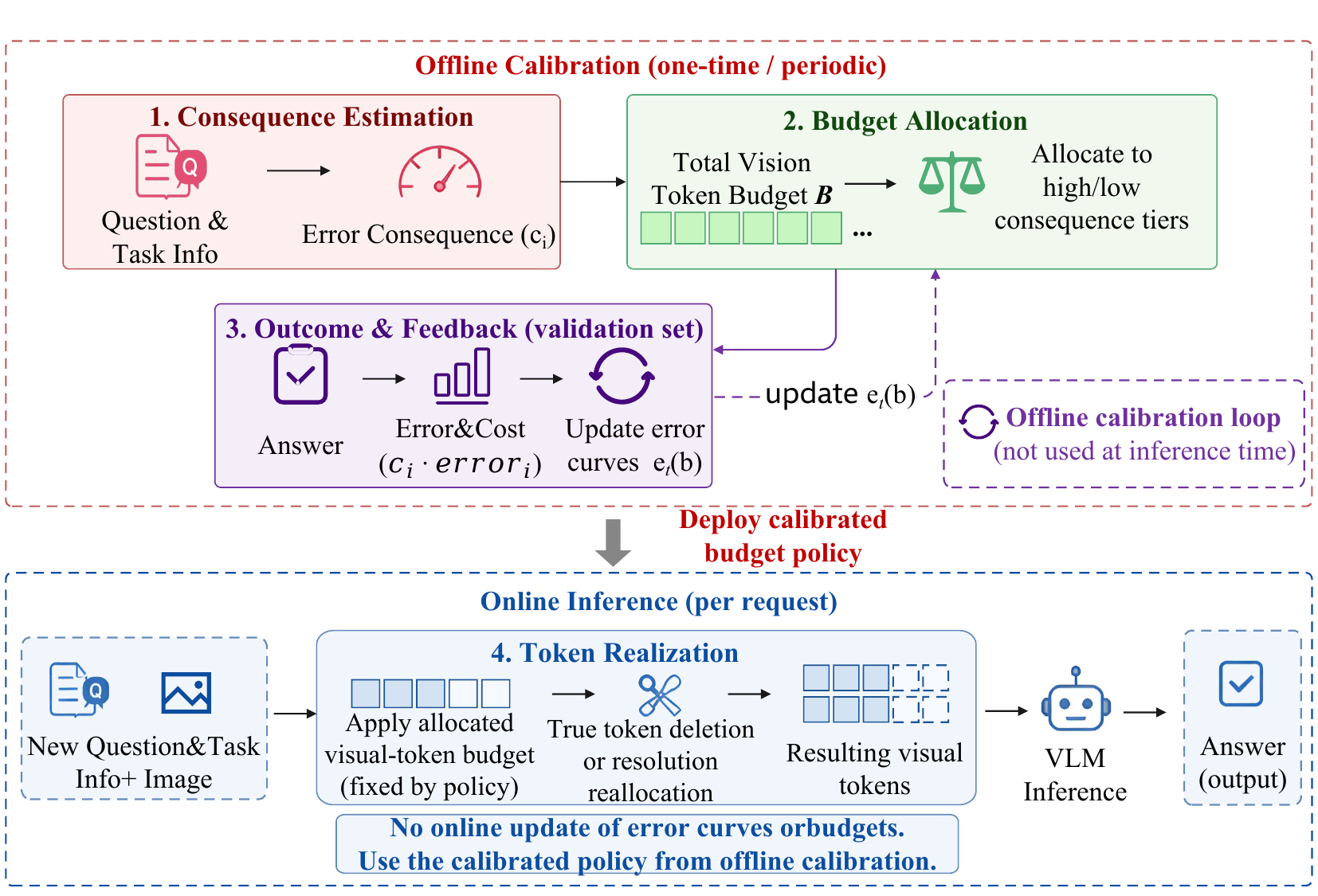}
\caption{Overview of consequence-sensitive visual token compression. During offline calibration, validation outcomes are used to estimate tier-specific error--budget curves 
and select calibrated budgets for a given total token budget. During online inference, each request is assigned to a consequence tier from question or task information, 
and the calibrated budget is applied before VLM inference. 
The feedback loop updates error--budget curves only during offline calibration.}
\label{fig:framework}
\end{figure}

Figure~\ref{fig:framework} summarizes the proposed framework. 
The method consists of an offline calibration stage and an online inference stage. During the offline stage, consequence signals and validation outcomes are used to estimate 
tier-specific error--budget curves and select a calibrated budget policy. During the online stage, the policy is fixed: each request is assigned to a consequence tier using 
inference-time-observable question or task information, and the corresponding 
visual-token budget is realized before VLM inference.

\subsection{Problem}
Each sample $i$ carries a consequence weight $c_i > 0$, the cost incurred if the model answers it incorrectly, and receives a vision-token budget $b_i$. Samples are grouped into tiers $t \in \mathcal{T}$ with common weight $c_t$, tier size $n_t$, and a tier-level error-vs-budget curve $e_t(b)$, measured with the token selector held fixed. Given a hard total budget $B$, we seek
\begin{equation}
\min_{\{b_t\}} \; \sum_t n_t\, c_t\, e_t(b_t) \quad \text{s.t.} \quad \sum_t n_t b_t \le B,\;\; b_t \ge b_{\min},
\end{equation}
where $b_{\min}$ is the smallest measured budget (a grid edge, not a known system floor). In our two-tier instantiation ($n_{\mathrm{hi}} = n_{\mathrm{lo}} = n$, $c_{\mathrm{hi}} = r$, $c_{\mathrm{lo}} = 1$, $B_{\mathrm{hi}} + B_{\mathrm{lo}} = 608$, i.e., 304 tokens/image on average), the objective equals $n(r{+}1)$ times the cost-weighted error
\begin{equation}
J(B_{\mathrm{hi}}) = \frac{r\, E_{\mathrm{hi}}(B_{\mathrm{hi}}) + E_{\mathrm{lo}}(608 - B_{\mathrm{hi}})}{r+1},
\end{equation}
so minimizing either is equivalent.

\subsection{Allocation Principle}
For a continuous, differentiable relaxation, the Lagrangian $\mathcal{L} = \sum_t n_t c_t e_t(b_t) + \lambda(\sum_t n_t b_t - B) - \sum_t \mu_t n_t (b_t - b_{\min})$ with $\lambda, \mu_t \ge 0$ gives $c_t e_t'(b_t) = \mu_t - \lambda$; since every $e_t$ is decreasing, $\lambda > 0$ and the budget constraint is tight, and for tiers with inactive floor ($\mu_t = 0$) the cost-weighted marginal error reduction $c_t |e_t'(b_t)| = \lambda$ is equalized---the classical water-filling rule in the convex case. Our measured problem violates both premises of that rule: budgets live on a discrete 13-point grid, which invalidates derivative-based conditions outright, and the curves are non-convex (finite-difference marginals of $E_{\mathrm{hi}}$ are non-monotone: $-0.043$, $-0.023$, $-0.064$ per 32 tokens across $32{\to}64{\to}96{\to}128$), so even on the relaxation stationarity does not imply local minimality. The appropriate discrete first-order condition is exchange-based: at a grid optimum, moving one grid step of budget from either tier to the other does not decrease $J$. Our allocator therefore minimizes the measured $J$ by exhaustive search over the feasible pairs; the swap condition holds at every reported argmin by construction, and the continuous marginal condition is used only descriptively, where it happens to hold.

\subsection{Calibrate-Then-Allocate}
\label{sec:calibrate}
The allocation procedure follows a calibrate-then-allocate strategy. 
First, we estimate the tier-specific error--budget curves $\hat{e}_t(b)$ on a calibration split by evaluating each candidate budget in the predefined grid. 
We then select the optimal allocation by minimizing the estimated cost-weighted objective:

\begin{equation}
(B_{\mathrm{hi}}, B_{\mathrm{lo}})
=
\arg\min_{B_{\mathrm{hi}},B_{\mathrm{lo}}}
\hat{J}(B_{\mathrm{hi}},B_{\mathrm{lo}}),
\label{eq:calibration-allocation}
\end{equation}
subject to the fixed total budget constraint. 
During inference, each query is assigned to a consequence tier using an inference-time-observable signal. 
In our experiments, this signal is obtained either from a frozen keyword rule over the question text (e.g., amounts, dates, counts, and identifiers $\Rightarrow$ high consequence; descriptive queries $\Rightarrow$ low consequence) or from task identity in the mixed-workload setting. 
The corresponding tier budget is then applied to the input.

The calibration stage requires $O(|\mathrm{grid}| \times |\mathrm{calibration\ set}|)$ offline forward passes, while online inference overhead is limited to lightweight tier assignment.

\subsection{Two Allocation Mechanisms}
We adopt absolute token transfer rather than proportional allocation because preliminary analysis shows proportional shifting suffers from token dilution in dense visual inputs (detailed in Technical Supplement Section~2: Proportional Allocation and Token Dilution).

A per-image budget can be realized by (a) true token deletion: encode at full resolution, rank tokens with a fixed selector, and delete the rest exactly (shortened embeddings and position ids; zero-masking instead of deletion produces out-of-distribution zero vectors and collapses generation); or (b) resolution reallocation: resize the image so the encoder emits approximately the budgeted token count, then run the standard generate path unmodified. Mechanism (b) requires no model-specific code, applies to any VLM, and---as we show---dominates (a) on document images for Qwen2.5-VL; we therefore treat (b) as the deployment mechanism and (a) as the analysis instrument that additionally enables selector-level control.

\subsection{A Two-Axis Map of Allocation Regimes}
\label{sec:map}
Question classes can differ along two orthogonal axes: a sensitivity gap (do their marginal error reductions per token differ?) and a cost gap (do their errors cost differently?). In cross-task benchmarks (DocVQA vs.\ synthetic solid-color images), there is a sensitivity gap and no cost gap: a content-driven allocator, reading only the image, routes tokens to the sensitive class and wins; cost labels carry no additional information by definition. Our within-task benchmark works differently by design: both classes see the same dense-document images and their measured token sensitivities are statistically indistinguishable across the budget grid, so an image-only allocator is blind by design; a cost-blind allocator is also definitionally handicapped on a cost-weighted metric, so the empirical content of that comparison is the gap's magnitude, not its sign. When neither gap exists (saturated synthetic images) no allocator helps; the both-gaps cell is untested in this paper, and composition there is a hypothesis, not a result. In the regimes we test, consequence-sensitive allocation is necessary exactly when a cost gap exists without a sensitivity gap that image content could proxy.

\section{Experimental Setup}
\label{sec:setup}

\paragraph{Model and mechanisms.}
Qwen2.5-VL-7B \citep{qwen25vl} (bf16, A100) is the primary model, with LLaVA-OneVision-7B \citep{llavaov} as the second architecture. Unless stated otherwise, budgets are realized by true deletion with a fixed redundancy selector (keep the most mutually diverse tokens); the resolution mechanism and alternative selectors are evaluated in the invariance study. Greedy decoding, $\le$16 new tokens, questions suffixed with "Answer briefly."

\paragraph{Benchmarks.}
Within-task (main): DocVQA validation \citep{docvqa}, images resized to max side 1008 ($\ge$640 native tokens), split by a frozen keyword rule on the question text into 300 transactional (high, $c{=}5$) and 300 descriptive (low, $c{=}1$) questions, drawn disjointly from a 120-sample pilot. Cross-task (diagnostic): 60 DocVQA questions vs.\ 60 color-identification questions on $1008^2$ solid-color images (1296 native tokens each, so every strategy's budget is feasible). Generalization: ChartQA test \citep{chartqa} and InfographicVQA validation \citep{infovqa}, 200/200 each, same frozen rule verbatim. Mixed workload: 120 VQAv2 \citep{vqav2} (low, $c{=}1$), 120 TextVQA \citep{textvqa} (medium, $c{=}3$), 120 DocVQA (high, $c{=}5$), tiers given by task identity.

\paragraph{Equal-budget protocol.}
The cost-aware assignment fixes the total budget $B$; uniform gives every image $B/N$; anti reverses the tiers (direction control); content allocates $B$ proportionally to each image's feature-level information density (1 $-$ mean pairwise cosine similarity), clamped and redistributed to preserve $B$ exactly. Every run asserts, per image, that the assigned budget does not exceed the native token count (no silent clamping) and, per strategy, that realized totals are exactly equal (deviations for the discrete-resolution mechanisms are reported: $\le$0.4\% for resolution, $\le$0.02\% for LLaVA rung alignment).

\paragraph{Statistics.}
Per benchmark, the pre-registered primary comparison is high-tier error, uniform vs.\ cost-aware, by exact McNemar on paired correctness; $\Delta_{\mathrm{CWE}}$ is assessed by paired bootstrap (10k resamples) with the same weights used for allocation; answers are graded by a normalized containment match shared by all arms. We report discordant counts $b{:}c$ (uniform-wrong/cost-right vs.\ converse) throughout. Additional implementation details, including dataset construction, budget verification, and statistical protocols, are provided in the Technical Supplement, Section~1 
(Full Experimental Setup and Implementation Details).

\section{Results}
\label{sec:results}

\subsection{Attribution: When Do Cost Labels Matter?}
\label{sec:attribution}

\paragraph{Cross-Task Analysis: Separating Content Effects from Consequence Effects.}

Table~\ref{tab:crosstask} and Table~\ref{tab:main} realize the two tested cells of the map in Section~\ref{sec:map}. 
We first examine a cross-task setting where content information can reveal token sensitivity differences. 
In this setting, the content allocator squeezes saturated solid-color images to the floor and outperforms even our cost-aware assignment, showing that content signals alone are sufficient when visual sensitivity differs across tasks. 
Therefore, gains observed in such settings cannot be attributed to consequence information, motivating a controlled within-task evaluation. Additional task-level sensitivity diagnostics and failure cases are analyzed in
the Technical Supplement, Section~6 (Task-Level Sensitivity Diagnostics).

\paragraph{Within-Task Evaluation: Isolating Consequence-Aware Allocation.}

To isolate the effect of consequence information, we construct a within-task setting where high- and low-consequence questions share the same document image distribution. 
Under this setting, the content allocator is blind to consequence differences because visual inputs are identical across tiers. 
It therefore assigns nearly uniform budgets and becomes statistically indistinguishable from uniform allocation, while consequence-aware allocation converts 54 high-stakes errors against only 4 reversals. 

The trade-off introduced by consequence-aware allocation is explicit: descriptive-tier error increases to 0.843, resulting in worse unweighted accuracy (0.488 vs.\ 0.337 for uniform). 
This demonstrates that the method is optimizing the intended cost-weighted objective rather than simply recovering an accuracy-maximizing allocation.

Two further checks support this attribution. 
First, the two tiers' error-vs-budget curves are statistically indistinguishable in slope across the 13-point grid (central-difference marginals at the uniform split: $0.00173$ vs.\ $0.00171$, slope SE ${\approx}2{\times}10^{-4}$), indicating that the observed improvement is not caused by high-consequence samples being inherently more token-sensitive. 
Second, our earlier proportional-budget benchmarks primarily reflected token sensitivity differences; the absolute token-transfer design adopted here removes this confound and isolates the contribution of consequence information. For attribution, Table~\ref{tab:main} uses a pre-specified absolute-transfer 
allocation rather than the post-hoc grid optimum; the calibrated optimum is 
studied separately in Section~\ref{sec:frontier}.
\begin{table}[t]
\centering
\small
\begin{tabular}{lccc}
\toprule
Strategy & High err. & Low err. & CWE ($r{=}5$) \\
\midrule
uniform (304/img) & 0.400 & 0.000 & 0.333 \\
content & \textbf{0.050} & 0.000 & \textbf{0.042} \\
cost-aware (512/96) & 0.117 & 0.000 & 0.097 \\
anti & 0.783 & 0.000 & 0.653 \\
\bottomrule
\end{tabular}
\caption{\textbf{Cross-task diagnostic} (DocVQA vs.\ solid-color, $N{=}120$, equal realized budget 36{,}480). With a sensitivity gap and no real cost gap, the content allocator wins without cost labels---this setting cannot attribute gains to consequence.}
\label{tab:crosstask}
\end{table}

\begin{table}[t]
\centering
\small
\adjustbox{max width=\columnwidth}{
\begin{tabular}{lccccc}
\toprule
Strategy & High & Low & Avg & CWE & McNemar \\
 & err. & err. & err. & ($r{=}5$) & ($b{:}c$, $p$) \\
\midrule
uniform & 0.300 & 0.373 & 0.337 & 0.312 & --- \\
content & 0.300 & 0.383 & 0.342 & 0.314 & 4:4, $1.0$ \\
cost-aware & \textbf{0.133} & 0.843 & 0.488 & \textbf{0.252} & 54:4, ${\approx}3{\times}10^{-12}$ \\
anti & 0.837 & 0.130 & 0.484 & 0.719 & 2:163 \\
\bottomrule
\end{tabular}
}
\caption{\textbf{Within-task attribution result} 
(DocVQA, $N{=}600$, equal realized budget 182{,}400, pre-specified budgets (512,\,96)). 
The image distribution is identical across tiers: the content allocator assigns 
${\approx}$uniform budgets (305/303) and performs like uniform, while consequence-sensitive 
allocation cuts high-stakes error $2.25\times$. 
Avg err.\ (unweighted, equal tier sizes) rises for cost-aware (0.488 vs.\ 0.337), 
showing that the method optimizes cost-weighted error rather than average accuracy. 
The calibrated optimum under $r{=}5$ is analyzed separately in the allocation frontier.}
\label{tab:main}
\end{table}
\paragraph{Do the tiers track real consequence?} The frozen rule is a keyword proxy, so we validate it two ways on the confirmatory 600. Additional validation of the consequence signals is provided in the Technical Supplement, Section~7
(Validation of Consequence Signals). Against gold-answer content, $90.3\%$ of high-tier questions have a quantitative gold answer (amount, date, count, code) versus $32\%$ of low-tier ones ($\kappa{=}0.58$); against an independent frontier-LLM consequence judge, blind to the rule, on a stratified 200-question sample, agreement is $72.5\%$ ($\kappa{=}0.45$). Disagreements are systematic: the rule over-labels navigational numbers (page/table/chapter) as high and, lacking the keywords budget/tel/zip, leaks some genuinely transactional questions into low---which makes the reported effect conservative, as high-value questions are then under-served. Deployment does not require the hand-written rule: a lightweight classifier predicts the tier from question text with $94\%$ 5-fold accuracy and better matches the judge's consequence labels than the rule itself. Additional details are provided in the Technical Supplement, Section~9 (Automatic Consequence Signal Prediction).

\subsection{The Allocation Frontier}
\label{sec:frontier}
\begin{figure*}[t]
\centering
\includegraphics[width=0.9\textwidth]{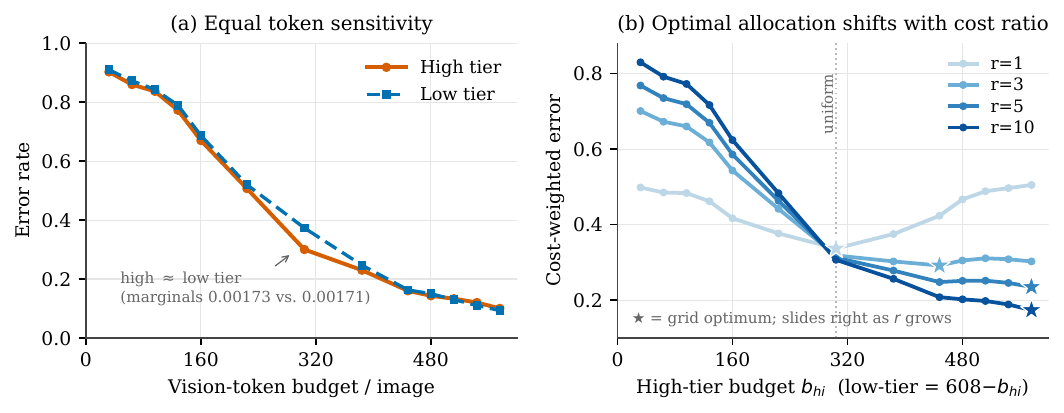}
\caption{\textbf{The allocation frontier} (within-task DocVQA, $N{=}600$). 
\textbf{(a)} High- and low-consequence questions have nearly identical 
error--budget curves, showing that the allocation gain is not caused by 
different token sensitivity. 
\textbf{(b)} Cost-weighted error as a function of the high-tier budget 
$B_{\mathrm{hi}}$ under different consequence ratios with a fixed total budget 
of 304 tokens per image on average. 
The optimum remains uniform at $r{=}1$ and shifts toward high-consequence 
samples as the consequence ratio increases.}
\label{fig:frontier}
\end{figure*}
\paragraph{Token-Response Curves (Equal Sensitivity).}
\label{sec:curves}
\begin{table}[t]
\centering
\small
\adjustbox{max width=0.9\columnwidth}{%
\begin{tabular}{lcc cccc}
\toprule
$(B_{\mathrm{hi}},B_{\mathrm{lo}})$ & $E_{\mathrm{hi}}$ & $E_{\mathrm{lo}}$ & \multicolumn{4}{c}{CWE at cost ratio $r$} \\
\cmidrule(lr){4-7}
 & & & $r{=}1$ & $r{=}3$ & $r{=}5$ & $r{=}10$ \\
\midrule
(304, 304) & 0.300 & 0.373 & \textbf{0.337} & 0.318 & 0.312 & 0.307 \\
(384, 224) & 0.230 & 0.520 & 0.375 & 0.302 & 0.278 & 0.256 \\
(448, 160) & 0.160 & 0.687 & 0.423 & \textbf{0.292} & 0.248 & 0.208 \\
(480, 128) & 0.143 & 0.790 & 0.467 & 0.305 & 0.251 & 0.202 \\
(512, 96)  & 0.133 & 0.843 & 0.488 & 0.311 & 0.252 & 0.198 \\
(544, 64)  & 0.120 & 0.873 & 0.497 & 0.308 & 0.246 & 0.188 \\
(576, 32)  & 0.100 & 0.910 & 0.505 & 0.302 & \textbf{0.235} & \textbf{0.174} \\
\bottomrule
\end{tabular}
}
\caption{\textbf{Measured frontier} (within-task DocVQA, $N{=}600$; all allocations use the same total budget). Bold values indicate the grid optimum under each consequence ratio: uniform allocation for $r\le2$, an interior optimum for $r=3$, and boundary solutions when the low-tier budget becomes limited.}
\label{tab:frontier}
\end{table}
Figure~\ref{fig:frontier}(a) visualizes the tier-specific error--budget curves on the confirmatory set; the full tabulated values are reported in the Technical Supplement, Section~3 (Additional Allocation Frontier Details). The two tiers' error rates remain nearly identical across the entire budget range; the central-difference marginals at the uniform split, computed from the tabulated values, are $0.00173$ for high-consequence samples and $0.00171$ for low-consequence samples, with a slope standard error of approximately $2\times10^{-4}$. Therefore, high- and low-consequence questions exhibit statistically indistinguishable token sensitivity.

This result isolates the source of the allocation gain. The improvement is not caused by high-consequence samples being inherently more sensitive to additional visual tokens. Under symmetric cost ($r=1$), where both tiers have equal importance, the optimal allocation remains uniform. Only when consequence asymmetry is introduced does the optimal allocation shift toward high-consequence samples.

\paragraph{Allocation Frontier.}

We next characterize how the optimal allocation changes as consequence asymmetry increases. Figure~\ref{fig:frontier} visualizes the allocation frontier, and Table~\ref{tab:frontier} reports the measured cost-weighted error for different high-to-low consequence ratios while keeping the total token budget fixed.

When the consequence ratio is symmetric ($r=1$), uniform allocation is optimal because the two tiers exhibit equivalent token-response behavior. As the consequence ratio increases, the optimal allocation progressively shifts tokens toward high-consequence samples. At moderate asymmetry ($r=3$), the optimal solution lies inside the allocation space, where the weighted benefits of transferring additional tokens are approximately balanced. When the consequence ratio becomes larger ($r\ge5$), the optimal allocation moves toward the boundary of the evaluated budget grid, indicating that additional computation should continue to be transferred toward high-consequence samples.

The anti-direction allocation provides a control condition. 
Moving tokens away from high-consequence samples consistently increases 
cost-weighted error in the frontier (Figure~\ref{fig:frontier}(b)). 
At the extreme anti-direction point, the high-tier budget is reduced to 
$B_{\mathrm{hi}}{=}32$, where the high-tier error reaches $0.903$ 
(Figure~\ref{fig:frontier}(a)). 
These results show that the allocation frontier is determined by consequence 
asymmetry rather than average accuracy optimization.

The measured frontier is obtained directly from empirical error--budget curves 
and therefore represents a discrete allocation boundary rather than an analytical 
optimum. We further examine whether the selected allocations generalize beyond 
the calibration samples.

\begin{table}[t]
\centering
\small
\adjustbox{max width=\columnwidth}{%
\begin{tabular}{lccccc}
\toprule
Benchmark & $N$ & \multicolumn{2}{c}{High-tier err.} & McNemar & Break-even \\
\cmidrule(lr){3-4}
 & & unif. & cost & ($b{:}c$, $p$) & $r$ \\
\midrule
InfoVQA & 400 & 0.660 & 0.410 & 57:7, ${<}10^{-9}$ & 0.7 \\
DocVQA & 600 & 0.300 & 0.133 & 54:4, $3{\times}10^{-12}$ & 2.8 \\
ChartQA & 400 & 0.335 & 0.235 & 22:2, $4{\times}10^{-5}$ & 3.6 \\
\bottomrule
\end{tabular}
}
\caption{\textbf{Cross-dataset replication with the tier rule frozen verbatim.} The absolute high-tier reduction is monotone in information density (0.250 / 0.167 / 0.100), and the break-even cost ratio falls as density rises.}
\label{tab:generalization}
\end{table}

Split-half calibration fits the two error-budget curves on odd-indexed samples and evaluates the predicted allocation on the even-indexed samples. The predicted optimum achieves rank 1/13 at $r\in\{1,10,20\}$ and rank 3/13 at $r\in\{2,5\}$ (held-out CWE gaps $+0.007$ and $+0.003$), but only rank 6/13 at $r=3$ (gap $+0.023$), where the optimum lies in the interior regime and is more sensitive to curve estimation errors.
The deployed optimum selected at $r=5$, $(576,32)$, improves high-tier error over uniform allocation with discordant pairs $65{:}5$ (exact $p\approx2\times10^{-14}$) and achieves a $\Delta_{\mathrm{CWE}}$ 95\% confidence interval of $[0.032,0.122]$.

\subsection{Generalization Across Dense Benchmarks}
\label{sec:generalization}

Table~\ref{tab:generalization} shows the same frozen question rule transferring across three benchmarks. The gradient is the applicability boundary made quantitative: the denser the task, the larger the absolute gain and the lower the cost asymmetry needed to justify transfer---to the point that on InfographicVQA reallocation wins even on plain accuracy (break-even $0.7 < 1$). We state the CWE-level statistics plainly: at $r{=}5$ the paired bootstrap CI on $\Delta_{\mathrm{CWE}}$ excludes zero for DocVQA ($[0.018, 0.102]$) and InfographicVQA ($[0.117, 0.238]$) but not for ChartQA ($[-0.019, 0.065]$), whose break-even (3.6) sits close to $r{=}5$; ChartQA's allocation direction is nonetheless decisively significant (22:2), and its CWE margin grows with $r$. The high-tier reduction is not an artifact of the lenient scorer: it persists on all three datasets under strict exact match and the official DocVQA ANLS@$0.5$ metric; see the Technical Supplement, Section~8 (Metric Robustness), e.g., DocVQA McNemar $64{:}5$ and $44{:}3$. Full cross-dataset results, including all allocation strategies and detailed
statistical comparisons, are reported in the Technical Supplement, Section~4
(Full Cross-Dataset Results).
\subsection{Invariance: Mechanism, Model, Selector}
\label{sec:invariance}

\begin{table}[t]
\centering
\small
\begin{tabular}{llccc}
\toprule
Axis & Variant & \multicolumn{2}{c}{High-tier err.} & McNemar \\
\cmidrule(lr){3-4}
 & & unif. & cost & ($b{:}c$, $p$) \\
\midrule
Mechanism & resolution & 0.250 & 0.123 & 47:9, ${<}10^{-6}$ \\
Model & LLaVA-OV & 0.417 & 0.330 & 40:14, $5{\times}10^{-4}$ \\
Selector & random & 0.497 & 0.247 & 80:5, $2{\times}10^{-18}$ \\
Selector & saliency & 0.377 & 0.127 & 80:5, $2{\times}10^{-18}$ \\
Selector & FastV & 0.560 & 0.297 & 84:5, ${<}10^{-18}$ \\
\bottomrule
\end{tabular}
\caption{\textbf{Invariance study} (within-task DocVQA, same 600 samples). Rows vary one axis at a time from the main setting (deletion / Qwen2.5-VL / redundancy).}
\label{tab:invariance}
\end{table}

\paragraph{Mechanism.} 
Table~\ref{tab:invariance} summarizes the invariance results across
different budget realization mechanisms, model architectures, and token selectors.
Realizing budgets by resolution instead of deletion improves every arm 
(uniform 0.250 vs.\ 0.300)---consistent with Qwen2.5-VL's native-dynamic-resolution training---
while preserving the allocation gain (47:9). The consequence-sensitive gain is a property of the allocation, not of any one token-reduction mechanism, and its strongest instantiation requires no model surgery at all.

\paragraph{Model.} LLaVA-OneVision shares no encoder, positional scheme, or tiling strategy with Qwen2.5-VL, and its AnyRes ladder quantizes budgets to coarse rungs (aligned to equal totals within 0.02\%). The ordering replicates (anti: 14:90, $p\!\approx\!9\times10^{-15}$), with a smaller absolute gain (its minimum rung already spends 1306 tokens), and a break-even ratio of 2.8---nearly identical to Qwen's 2.8, an invariance we report as an observation, not a claim. As with ChartQA, the $r{=}5$ CWE margin is directionally positive but its CI includes zero ($[-0.009, 0.073]$).

\paragraph{Selector Robustness.}
\label{sec:selector}

To examine whether consequence-sensitive allocation depends on a specific token selector,
we evaluate the same allocation policy with different visual token selection strategies,
including random selection, saliency ranking, FastV~\citep{fastv},
SparseVLM~\citep{sparsevlm}, and VisionZip~\citep{visionzip}.
Table~\ref{tab:selfull} summarizes the complete selector-level results.
The allocation gain consistently persists across all selection mechanisms. 
Random and saliency provide simple selection controls, while FastV, SparseVLM, and VisionZip represent stronger training-free pruning
strategies based on attention-guided pruning, text-guided relevance estimation,
and informative token selection, respectively.
Although these selectors produce different absolute error levels under uniform allocation,
consequence-sensitive allocation consistently reduces high-consequence errors under the
same total token budget.

These results show that consequence-sensitive allocation operates at the budget allocation
level and is orthogonal to the token selection mechanism.
Therefore, existing token-selection methods such as FastV, SparseVLM, and VisionZip
can be viewed as composable components rather than
competing alternatives.

\begin{table}[t]
\centering\small
\begin{tabular}{llcccc}
\toprule
Selector & Arm & High & Low & CWE$_{r5}$ & McNemar \\
\midrule
\multirow{2}{*}{redundancy} & unif. & 0.300 & 0.373 & 0.312 & --- \\
 & cost & \textbf{0.133} & 0.843 & \textbf{0.252} & $54{:}4$ \\
\midrule
\multirow{2}{*}{random} & unif. & 0.497 & 0.470 & 0.492 & --- \\
 & cost & \textbf{0.247} & 0.753 & \textbf{0.331} & $80{:}5$ \\
\midrule
\multirow{2}{*}{saliency} & unif. & 0.377 & 0.457 & 0.390 & --- \\
 & cost & \textbf{0.127} & 0.840 & \textbf{0.246} & $80{:}5$ \\
\midrule
\multirow{2}{*}{FastV} & unif. & 0.560 & 0.450 & 0.542 & --- \\
 & cost & \textbf{0.297} & 0.710 & \textbf{0.366} & $84{:}5$ \\ \midrule
 \multirow{2}{*}{SparseVLM} & unif. & 0.460 & 0.373 & 0.446 & --- \\
 & cost & \textbf{0.257} & 0.657 & \textbf{0.323} & $67{:}6$ \\
\midrule
\multirow{2}{*}{VisionZip} & unif. & 0.397 & 0.353 & 0.389 & --- \\
 & cost & \textbf{0.253} & 0.607 & \textbf{0.312} & $45{:}2$ \\
\bottomrule
\end{tabular}
\caption{\textbf{Full selector robustness} (within-task DocVQA, $N{=}600$).
Consequence-sensitive allocation remains effective across diverse token selection strategies,
including redundancy-based selection, random selection, saliency ranking,
FastV, SparseVLM, and VisionZip.
}
\label{tab:selfull}
\end{table}

\subsection{Mixed-Workload Deployment}
\label{sec:mixed}

\begin{table}[t]
\centering
\small
\begin{tabular}{lcccc}
\toprule
Strategy & High & Med & Low & CWE (5/3/1) \\
\midrule
uniform (288/img) & 0.475 & 0.183 & 0.183 & 0.345 \\
content & 0.500 & 0.175 & 0.167 & 0.355 \\
tiered (512/256/96) & \textbf{0.192} & 0.200 & 0.350 & \textbf{0.212} \\
anti & 0.825 & 0.200 & 0.125 & 0.539 \\
\bottomrule
\end{tabular}
\caption{\textbf{Three-tier mixed workload} (DocVQA / TextVQA / VQAv2, $N{=}360$, equal total budget). Tiers come from task identity---metadata that is free at deployment. McNemar (high tier): tiered 36:2 ($p{\approx}5{\times}10^{-9}$); content 2:5 ($p{=}0.45$).}
\label{tab:mixed}
\end{table}

On a realistic three-task mixture (Table~\ref{tab:mixed}), tier-based allocation cuts CWE by 38\% ($\Delta_{\mathrm{CWE}}$ CI $[0.080, 0.188]$). The content allocator fails informatively: real photographs, scene-text images, and documents have similar feature-level diversity, so its allocation collapses to near-uniform (273/291/299 tokens per tier)---the clean separation it enjoyed against synthetic solid colors (Table~\ref{tab:crosstask}) does not survive contact with real workloads, while task identity remains free and effective.

\subsection{Efficiency}
\label{sec:efficiency}

Using the resolution-based deployment mechanism, reducing the average visual tokens from full resolution to 304 tokens lowers latency from 482.1\,ms to 356.1\,ms ($-26.1\%$). The deployed two-tier allocation $(576,32)$ achieves about $21\%$ lower latency than full-resolution inference while improving CWE from 0.312 to 0.235. True deletion is used only as an analysis mechanism and is not itself an accelerator because it extracts full-resolution embeddings before deletion. Full latency measurements are reported in the Technical Supplement, Section~10 (Full Latency Sweep).

\section{Conclusion}
We formulate visual token compression as cost-weighted error minimization under a strict token budget, introduce a controlled evaluation that disentangles consequence information from content-derived signals, and characterize when consequence-sensitive allocation is beneficial. The resulting allocation frontier shows that uniform allocation is optimal when error costs are similar, whereas increasingly asymmetric costs justify shifting more visual computation toward high-consequence requests. On DocVQA, this transition occurs at a consequence ratio of approximately $2.8{:}1$, with break-even ratios ranging from $0.7$ to $3.6$ across benchmarks of different information density. The framework is lightweight and deployment-oriented, inferring consequence tiers from question or task information, calibrating tier-specific error--budget curves offline, and realizing the resulting budgets through token deletion or resolution reallocation. More broadly, our results suggest that visual token compression should be evaluated not only by how much accuracy it preserves, but also by which errors it prevents.

\bibliography{aaai2027}

\appendix
\section*{Technical Supplement}
This Technical Supplement provides additional experimental details, analyses, 
and robustness studies supporting the main paper 
"Not All Visual Tokens Are Equally Safe to Remove: Consequence-Sensitive Visual Token Compression."

The Full Experimental Setup and Implementation Details section describes 
the complete experimental setup, including model configurations, dataset construction, 
consequence assignment rules, budget constraints, scoring, and statistical protocols.

The Proportional Allocation and Token Dilution section analyzes why earlier 
ratio-based allocation was insufficient and motivates the absolute token-transfer 
design used in the main paper.

The Additional Allocation Frontier Details section reports the full token-response 
curves underlying the allocation frontier and supports the claim that the within-task 
high- and low-consequence tiers have nearly identical token sensitivity.

The Full Cross-Dataset Results section provides comprehensive per-arm evaluations 
on dense vision-language benchmarks, validating that the proposed allocation 
principle generalizes beyond the primary DocVQA setting.

The Additional Model and Scale Generalization section further evaluates the method 
across different model architectures and scales, including LLaVA-OneVision and 
Qwen2.5-VL-3B.

The Task-Level Sensitivity Diagnostics section presents task-level sensitivity 
analyses that motivate the choice of dense visual understanding tasks and clarify 
when consequence information provides additional value beyond content-based allocation.

The Validation of Consequence Signals section validates the consequence signals 
used for allocation through both objective answer-based analysis and independent 
LLM-based judgment.

The Metric Robustness section verifies that the observed improvements remain 
consistent across multiple evaluation metrics, including strict exact match and 
the official ANLS metric.

The Automatic Consequence Signal Prediction section demonstrates that consequence 
signals can be automatically predicted from question text using a lightweight 
classifier, supporting practical deployment without manually designed rules.

The Full Latency Sweep section reports detailed latency measurements for the 
resolution-based deployment mechanism. Finally, the Limitations section discusses 
the main assumptions and remaining scope of the study.

Together, these analyses provide additional evidence that consequence-sensitive 
visual token compression improves cost-weighted inference efficiency by allocating 
limited visual computation toward samples where errors are more costly.

\section{Full Experimental Setup and Implementation Details}
\label{app:setup}

\paragraph{Models and environment.} 
The primary model is Qwen2.5-VL-7B \citep{qwen25vl} in \texttt{bf16} on a single A100-80GB; the second model is LLaVA-OneVision-Qwen2-7B-ov-hf \citep{llavaov}. Decoding is greedy (\texttt{do\_sample=False}, argmax), $\le 16$ new tokens, with every question suffixed by ``\,Answer briefly.''

\paragraph{Datasets and disjoint sampling.} 
All samples are drawn by streaming. The within-task pilot ($N{=}120$) uses the first 257 scanned DocVQA validation items; the pre-registered confirmatory set ($N{=}600$) skips the first 257 items and scans 1184 items, so the pilot and confirmatory sets are strictly disjoint. ChartQA test scans 568 samples and InfographicVQA validation scans 601 samples from parquet files, supplying 200 high-consequence and 200 low-consequence samples each. The mixed workload uses 120 samples each from VQAv2, TextVQA, and DocVQA, with the DocVQA subset skipping the first 1184 scanned items. Images are resized to a maximum side of 1008 and filtered to have at least 640 native tokens so that every grid budget is feasible without clamping; InfographicVQA additionally up-scales extreme aspect ratios up to a 1512 maximum side.

\paragraph{Frozen cost-label rule.} 
A tier is assigned from the question text only, never from the image or answer, using a keyword rule frozen verbatim across all datasets. A question is assigned to the high-consequence tier ($c{=}5$) if it matches any of the following keywords: \{how much, how many, amount, total, number, no., date, when, year, price, cost, percent, percentage, \%, phone, fax, time, quantity, rate, value, figure, sum, id, code\}. Otherwise, it is assigned to the low-consequence tier ($c{=}1$). In the mixed workload, the tier is assigned by task identity: VQAv2 is low consequence, TextVQA is medium consequence ($c{=}3$), and DocVQA is high consequence.

\paragraph{Scoring.} 
The function \texttt{is\_correct} normalizes both prediction and reference strings by stripping whitespace, lowercasing, and dropping a trailing period. It accepts exact match or bidirectional substring containment. The identical scorer is shared by all arms.

\paragraph{Equal-budget protocol.} 
Each run asserts, per image, that the assigned budget does not exceed the native token count, preventing silent clamping. Each run also asserts, per arm, that realized totals are exactly equal; uniform allocation assigns $\lfloor B/N\rfloor$ tokens to every image, with the remainder given to the first samples. Realized-total deviations for discrete mechanisms are reported rather than hidden: they are at most $0.4\%$ for the resolution mechanism and at most $0.02\%$ for the LLaVA AnyRes ladder. Cost-weighted error is computed as
\[
\mathrm{CWE}
=
\frac{\sum_i c_i\,\mathbb{1}[\mathrm{wrong}_i]}
{\sum_i c_i}.
\]
The cost ratio $r\in\{1,2,3,5,10,20\}$ is applied by zero-GPU post-hoc reweighting of the same per-sample correctness vectors.

\paragraph{Statistics.} 
The pre-registered primary test is high-tier error, comparing uniform allocation against cost-aware allocation by two-sided exact McNemar test on discordant pairs $b{:}c$. The cost-weighted error difference $\Delta_{\mathrm{CWE}}$ is assessed using a paired bootstrap with 10k resamples and BCa intervals. Split-half calibration fits both tier-specific error--budget curves on odd-indexed samples and evaluates the predicted argmin on the even-indexed samples. Greedy decoding is deterministic, so the only randomness is bootstrap resampling and selector tie-breaking.

\section{Proportional Allocation and Token Dilution}
\label{app:ratio}

Our earlier design allocated budgets by a ratio: the mean keep-ratio was fixed at $0.5$, the high-side ratio was chosen from $\{0.55,0.65\}$, and the low-side ratio was solved to preserve the equal-budget constraint. This design produced only weak, non-significant gains. On 120 DocVQA-vs-color samples, cost-aware allocation achieved CWE $0.125$ versus uniform CWE $0.153$ ($1.22\times$). The larger $1.75\times$ gain measured at $N{=}48$ was small-sample optimism, corresponding to only a two-sample high-tier difference.

The main cause is token dilution. Dense document images carry a large native token count, so a proportional shift transfers only a small number of absolute tokens and the intended consequence-aware transfer is swamped. An early synthetic-image bug further compounded this issue: $336{\times}336$ images contain only 144 native tokens, so uniform and anti ratio budgets silently exceeded the native token count and broke the equal-budget protocol. These observations motivate the absolute token-transfer design used throughout the main paper.

\section{Additional Allocation Frontier Details}
\label{app:frontier}

Table~\ref{tab:curves} reports the full token-response curves used to construct the allocation frontier in the main paper. The high- and low-consequence tiers respond almost identically to additional visual tokens across the evaluated budget grid. This supports the within-task attribution claim that the observed allocation gain is driven by consequence asymmetry rather than by a difference in token sensitivity.

\begin{table}[t]
\centering\small
\begin{tabular}{cccccc}
\toprule
Budget & 32 & 64 & 96 & 128 & 160 \\
\midrule
High err. & 0.903 & 0.860 & 0.837 & 0.773 & 0.670 \\
Low err.  & 0.910 & 0.873 & 0.843 & 0.790 & 0.687 \\
\midrule
Budget & 224 & 304 & 448 & 512 & 576 \\
\midrule
High err. & 0.507 & 0.300 & 0.160 & 0.133 & 0.100 \\
Low err.  & 0.520 & 0.373 & 0.163 & 0.130 & 0.093 \\
\bottomrule
\end{tabular}
\caption{\textbf{Token-response curves} (within-task DocVQA, $N{=}600$). The two tiers respond to additional visual tokens almost identically.}
\label{tab:curves}
\end{table}

\section{Full Cross-Dataset Results}
\label{app:crossdata}

Table~\ref{tab:crossfull} reports all four arms on the three dense benchmarks. Content allocation is indistinguishable from uniform on all three datasets: DocVQA $4{:}4$, $p{=}1.0$; ChartQA $5{:}1$, $p{=}0.22$; and InfographicVQA $3{:}1$, $p{=}0.63$. Anti allocation is catastrophic everywhere. The break-even cost ratios are $2.82$ for DocVQA, $3.60$ for ChartQA, and $0.74$ for InfographicVQA. Paired-bootstrap $\Delta_{\mathrm{CWE}}$ 95\% confidence intervals at $r{=}5$ are $[0.018,0.102]$ for DocVQA and $[0.117,0.238]$ for InfographicVQA, both excluding zero; ChartQA has interval $[-0.019,0.065]$, which does not exclude zero because its break-even ratio $3.60$ sits close to $r{=}5$.

\begin{table}[t]
\centering\small
\begin{tabular}{llccc}
\toprule
Data & Strategy & High & Low & CWE$_{r5}$ \\
\midrule
\multirow{4}{*}{DocVQA} 
 & uniform & 0.300 & 0.373 & 0.312 \\
 & cost & \textbf{0.133} & 0.843 & \textbf{0.252} \\
 & anti & 0.837 & 0.130 & 0.719 \\
 & content & 0.300 & 0.383 & 0.314 \\
\midrule
\multirow{4}{*}{ChartQA} 
 & uniform & 0.335 & 0.365 & 0.340 \\
 & cost & \textbf{0.235} & 0.725 & \textbf{0.317} \\
 & anti & 0.650 & 0.275 & 0.588 \\
 & content & 0.315 & 0.370 & 0.324 \\
\midrule
\multirow{4}{*}{InfoVQA} 
 & uniform & 0.660 & 0.670 & 0.662 \\
 & cost & \textbf{0.410} & 0.855 & \textbf{0.484} \\
 & anti & 0.865 & 0.455 & 0.797 \\
 & content & 0.650 & 0.675 & 0.654 \\
\bottomrule
\end{tabular}
\caption{\textbf{Full cross-dataset results} ($N{=}600/400/400$; 300/300, 200/200, 200/200 high/low). McNemar tests for uniform vs.\ cost-aware allocation on the high tier: DocVQA $54{:}4$ ($3{\times}10^{-12}$), ChartQA $22{:}2$ ($4{\times}10^{-5}$), and InfographicVQA $57{:}7$ ($<10^{-9}$).}
\label{tab:crossfull}
\end{table}

\section{Additional Model and Scale Generalization}
\label{app:llava}

LLaVA-OneVision shares no encoder, positional scheme, or tiling strategy with Qwen2.5-VL. Its AnyRes ladder quantizes budgets to coarse rungs $\{1269,1793,{\sim}2900,4854,5589\}$, realized as low$\to$1306 tokens and high$\to$4732 tokens, with uniform and anti greedily aligned to equal totals within $0.02\%$. Table~\ref{tab:llavafull} reports the three arms. The ordering replicates: cost-aware allocation gives $40{:}14$, $p{\approx}5{\times}10^{-4}$, and anti allocation gives $14{:}90$, $p{\approx}9{\times}10^{-15}$. The absolute gain is smaller because the minimum LLaVA rung already spends 1306 tokens. The break-even ratio is $2.79$, near-identical to Qwen's $2.82$; we report this as an observation, not a claim. The $r{=}5$ $\Delta_{\mathrm{CWE}}$ confidence interval $[-0.009,0.073]$ includes zero, as with ChartQA.

\begin{table}[t]
\centering\small
\begin{tabular}{lcccc}
\toprule
Strategy & High & Low & Avg & CWE$_{r5}$ \\
\midrule
uniform & 0.417 & 0.440 & 0.428 & 0.421 \\
cost    & \textbf{0.330} & 0.683 & 0.507 & \textbf{0.389} \\
anti    & 0.670 & 0.383 & 0.527 & 0.622 \\
\bottomrule
\end{tabular}
\caption{\textbf{LLaVA-OneVision-7B} (within-task DocVQA, $N{=}600$, resolution mechanism).}
\label{tab:llavafull}
\end{table}

\paragraph{Smaller scale (Qwen2.5-VL-3B).} 
The effect also holds at a smaller model size. On Qwen2.5-VL-3B with the resolution mechanism on the same within-task 600 samples, the ordering replicates, as shown in Table~\ref{tab:q3bfull}. Cost-aware allocation cuts high-tier error from $0.277$ to $0.147$ with McNemar $49{:}10$, $p{\approx}3{\times}10^{-7}$; anti allocation yields $0{:}165$. The CWE decreases from $0.289$ to $0.256$ at $r{=}5$. The break-even ratio lies between $r{=}3$ and $r{=}5$, and the $r{=}5$ $\Delta_{\mathrm{CWE}}$ confidence interval $[-0.008,0.078]$ includes zero, as with ChartQA and LLaVA-OneVision. The gain therefore survives across model scale, from 3B to 7B, as well as across architecture.

\begin{table}[t]
\centering\small
\begin{tabular}{lcccc}
\toprule
Strategy & High & Low & Avg & CWE$_{r5}$ \\
\midrule
uniform & 0.277 & 0.353 & 0.315 & 0.289 \\
cost    & \textbf{0.147} & 0.800 & 0.473 & \textbf{0.256} \\
anti    & 0.827 & 0.187 & 0.507 & 0.720 \\
\bottomrule
\end{tabular}
\caption{\textbf{Qwen2.5-VL-3B} (within-task DocVQA, $N{=}600$, resolution mechanism; realized budgets within $0.4\%$).}
\label{tab:q3bfull}
\end{table}

\section{Task-Level Sensitivity Diagnostics}
\label{app:early}

Before the within-task design, we ran cross-task probes whose task confounds make them unsuitable as primary evidence but useful for motivating the density boundary. On synthetic solid-color images, which have low token sensitivity, budgeting is inert: true-deletion cost-aware allocation equals uniform allocation, with both achieving CWE $0.069$, because the task saturates by 25--50\% of tokens. On TextVQA, which shows weak sensitivity, cost-aware allocation did not help: CWE is $0.174$ for cost-aware allocation versus $0.139$ for uniform allocation, additionally confounded by unequal image sizes. Only on dense DocVQA does the premise that high-cost questions need more tokens hold clearly: full-budget high-tier error is $0.083$, rising to $0.542$ at 25\% tokens.

These diagnostics explain why the main evaluation uses within-task DocVQA and why the generalization set focuses on dense document-like tasks. The cross-task diagnostic in the main paper makes the complementary point: when a content signal can observe a sensitivity gap, it can win without any cost label. Therefore, only the within-task setting attributes the observed gain to consequence information rather than to visual token sensitivity differences.

\section{Validation of Consequence Signals}
\label{app:labelval}

The tier rule described in the Full Experimental Setup and Implementation Details section is a keyword proxy; we assess how well it tracks error consequence with two checks on the confirmatory 600 samples.

\paragraph{Objective check (gold-answer content).} 
A question's gold answer is labeled quantitative if any reference answer contains a digit or a currency/percent marker. Cross-tabulated against the tier, as shown in Table~\ref{tab:labelval}, $90.3\%$ of high-tier questions have a quantitative answer, indicating that the high tier is genuinely transactional. The low tier is noisier: $68\%$ are non-quantitative, while 96 transactional questions leak into the low tier. Agreement is $0.792$ with $\kappa{=}0.583$.

\paragraph{LLM-judge check (consequence rubric).} 
On a seed-fixed stratified sample of 200 questions, with 100 from each tier, an independent frontier LLM is shown only the question and is blind to the keyword rule. It rates each question under the following rubric: high consequence if a wrong answer is a specific monetary amount, date, count, rate, or formal record identifier, such as invoice, voucher, fund, serial, phone, fax, zip, or code, that propagates into a decision or record; low consequence if the answer is a name, title, type, heading, description, navigational locator such as a page/table/chapter number, or trivial schedule time. Agreement is $0.725$ with $\kappa{=}0.450$.

\paragraph{Disagreement taxonomy.} 
Both checks fail in the same two directions. First, in over-labeled high cases, keywords such as number match page, table, or chapter numbers and expressions such as ``the number at the bottom right,'' while time matches coffee-break or lunch schedule times. These are navigational or trivial values and account for approximately 60\% of the rule-high/judge-low cell. Second, in under-labeled low cases, questions whose answers are dollar amounts or formal contact fields fall into the low tier because their stems lack the listed keywords, for example ``the budget for...'', ``Tel no'', or ``zipcode''. Because these under-labeled cases place genuinely high-value questions in the low tier, the equal-budget experiment under-serves them. The measured cost-weighted gain is therefore a lower bound on what a cleaner consequence labeler could yield.

\begin{table}[t]
\centering\small
\begin{tabular}{lcccc}
\toprule
Check & Agree & $\kappa$ & \multicolumn{2}{c}{Confusion} \\
\cmidrule(lr){4-5}
 & & & H/H\,,\,L/L & H/L\,,\,L/H \\
\midrule
Gold-answer ($N{=}600$) & 0.792 & 0.583 & 271 , 204 & 29 , 96 \\
LLM judge ($N{=}200$) & 0.725 & 0.450 & 70 , 75 & 30 , 25 \\
\bottomrule
\end{tabular}
\caption{\textbf{Label validation.} H/H = rule-high and check-high, L/L = rule-low and check-low, H/L = rule-high but check-low, and L/H = rule-low but check-high. The high tier is clean; residual noise concentrates in the low tier and is conservative for the main claim.}
\label{tab:labelval}
\end{table}

\section{Metric Robustness}
\label{app:metric}

Our grader is a lenient normalized-containment match. To verify that the allocation gain is not an artifact of this scorer, we re-score the saved per-sample predictions, without re-inference, under two stricter metrics: strict exact match, which uses normalized equality with no substring containment, and the official DocVQA ANLS@$0.5$ metric, which thresholds average normalized Levenshtein similarity at $0.5$. Table~\ref{tab:metric} shows that the high-tier reduction and its significance survive on all three datasets under all three metrics. The reproduced relaxed column matches the main-text numbers exactly, confirming the re-scoring pipeline.

\begin{table}[t]
\centering\small
\begin{tabular}{llccc}
\toprule
Data & Metric & High (u$\to$c) & CWE (u$\to$c) & McN. \\
\midrule
\multirow{3}{*}{DocVQA} 
 & relaxed & 0.300$\to$0.133 & 0.312$\to$0.252 & 54:4 \\
 & exact & 0.360$\to$0.163 & 0.382$\to$0.288 & 64:5 \\
 & ANLS & 0.247$\to$0.110 & 0.259$\to$0.227 & 44:3 \\
\midrule
\multirow{3}{*}{ChartQA} 
 & relaxed & 0.335$\to$0.235 & 0.340$\to$0.317 & 22:2 \\
 & exact & 0.470$\to$0.370 & 0.468$\to$0.443 & 21:1 \\
 & ANLS & 0.405$\to$0.325 & 0.398$\to$0.397 & 17:1 \\
\midrule
\multirow{3}{*}{InfoVQA} 
 & relaxed & 0.660$\to$0.410 & 0.662$\to$0.484 & 57:7 \\
 & exact & 0.675$\to$0.445 & 0.677$\to$0.517 & 53:7 \\
 & ANLS & 0.570$\to$0.355 & 0.562$\to$0.421 & 52:9 \\
\bottomrule
\end{tabular}
\caption{\textbf{Metric robustness} (uniform$\to$cost-aware, high tier and CWE at $r{=}5$). All McNemar tests have $p<10^{-4}$. The direction and significance hold under strict exact match and official ANLS@$0.5$.}
\label{tab:metric}
\end{table}

\section{Automatic Consequence Signal Prediction}
\label{app:predictor}

The consequence tier is cheaply predictable from question text, so deployment need not rely on the hand-written keyword rule. We fit a lightweight TF-IDF uni/bi-gram logistic-regression classifier and evaluate it by 5-fold cross-validation, as shown in Table~\ref{tab:predictor}. When predicting the frozen rule's own tier, it reaches $94.3\%$ accuracy with $\kappa{=}0.887$. This means the rule is recovered almost exactly from text, so an online predictor can stand in for the keyword list. When predicting the independent LLM-judge consequence described in the Validation of Consequence Signals section, it reaches $83.5\%$ accuracy with $\kappa{=}0.672$, which is higher than the fixed rule's own $72.5\%$ accuracy and $\kappa{=}0.45$ agreement with the judge. This improvement occurs because a learned predictor generalizes to unlisted transactional cues such as budget, tel, and zip. The frontier-LLM judge described in the Validation of Consequence Signals section is itself a second, training-free predictor.

\begin{table}[t]
\centering\small
\begin{tabular}{lcccc}
\toprule
Prediction target & $N$ & Acc & F1 & $\kappa$ \\
\midrule
Frozen rule tier & 600 & 0.943 & 0.942 & 0.887 \\
LLM-judge consequence & 200 & 0.835 & 0.839 & 0.672 \\
\bottomrule
\end{tabular}
\caption{\textbf{Automatic consequence prediction} from question text (TF-IDF + logistic regression, 5-fold CV). A lightweight learned predictor recovers the deployed tier and, on the judge's labels, exceeds the fixed rule's agreement with the judge.}
\label{tab:predictor}
\end{table}

\section{Full Latency Sweep}
\label{app:latency}

Table~\ref{tab:wallfull} reports the full latency sweep for the resolution-based deployment mechanism. Latency decreases as the number of visual tokens is reduced, but the gain plateaus near 160 tokens because decoding and fixed overhead begin to dominate. Reducing visual tokens from full resolution to 304 tokens lowers mean latency from 482.1\,ms to 356.1\,ms, a 26.1\% reduction. The deployed two-tier point $(576,32)$ achieves approximately 21\% lower latency than full-resolution inference.

\begin{table}[t]
\centering\small
\begin{tabular}{lccc}
\toprule
Setting & Mean (ms) & Tokens & Latency $\downarrow$ \\
\midrule
full generate & 482.1 & 959 & --- \\
resolution 576 & 415.8 & 572 & 13.8\% \\
resolution 448 & 391.1 & 450 & 18.9\% \\
resolution 304 & 356.1 & 304 & 26.1\% \\
resolution 160 & 353.3 & 162 & 26.7\% \\
resolution 96  & 362.6 & 96  & 24.8\% \\
\bottomrule
\end{tabular}
\caption{\textbf{Full latency sweep} (resolution mechanism). Latency decreases with token reduction but plateaus near 160 tokens due to decoding and fixed computation overhead. The deployed two-tier point $(576,32)$ achieves approximately $21\%$ lower latency than full-resolution inference.}
\label{tab:wallfull}
\end{table}

\section{Limitations}
\label{sec:limitations}

Our study has several limitations. The consequence weights are predefined design parameters rather than measured downstream costs, and consequence categories are assumed to be available at inference time. Although we evaluate cost-ratio sweeps and show that lightweight predictors can approximate consequence labels, real deployment would require workflow-specific cost estimation and uncertainty-aware consequence prediction. The allocation frontier is estimated from discrete empirical error--budget curves, so the optimum depends on the evaluated grid and may not coincide with a continuous optimum. Finally, our evaluation focuses on dense vision-language workloads where consequence effects can be isolated from visual sensitivity; broader workloads with simultaneously varying difficulty, token sensitivity, and error cost remain open.

\end{document}